\documentclass{jfpda}

\usepackage{verbatim}
\usepackage{amsmath}
\usepackage{algorithm}
\usepackage[noend]{algpseudocode}

\usepackage{natbib}
\usepackage[english]{babel}
\usepackage[latin1]{inputenc}
\usepackage[T1]{fontenc}

\shorttitle{Learning Macro-Actions for State-Space Planning}

\shortouvrage{JFPDA 2016}

\title{Learning Macro-actions for State-Space Planning}

\author{Sandra Castellanos-Paez\inst{\textbf{*}}, Damien Pellier\inst{*}, Humbert Fiorino\inst{*}, Sylvie Pesty\inst{*}}

\institute{
*Univ. Grenoble Alpes, LIG, F-38000 Grenoble, France\\
\texttt{firstname.lastname@imag.fr}
}

\begin{document}
\maketitle

\begin{abstract}
\begin{sloppypar}
Planning has achieved significant progress in recent years. Among the various approaches to scale up plan synthesis, the use of macro-actions has been widely explored. 
As a first stage towards the development of a solution to learn on-line macro-actions, we propose an algorithm to identify useful macro-actions based on data mining techniques. 
The integration in the planning search of these learned macro-actions shows significant improvements over four classical planning benchmarks. \end{sloppypar}
  \motscles{Automated Planning, Data Mining, Macro-actions, Learning}
\end{abstract}

\section{Introduction}
\begin{sloppypar}
Automated planning comes up with the challenge of devising fast and powerful systems that can autonomously find a plan to achieve a set of goals. A plan is a structure of appropriate actions for a given problem over a domain. From a frequent sequence of some of these actions, it is possible to build a meta-action structure called \textit{macro-action}. Finding the way to learn macro-actions from previously acquired knowledge (plans) can allow us to go quickly deep into the search space by triggering them during the search. 
While most of the literature presents the macro-action synthesis as a result of using a filter approach over previously detected macro-actions in a set of training problems \cite{Botea:macro-ff,Newton:2007,Jonsson:2009,Dulac}, we propose a novel and general macro learning method as part of an effort to detect automatically potential macro-actions and use them in the planning search.
\end{sloppypar}

\section{Mining useful Macro-actions}
\begin{sloppypar}

The main idea of our approach is to build macro-actions from sequential patterns of actions in a set of plans and to use them during the planning search to improve the performance of the planning system. 

A \textit{sequential pattern of actions} is a frequent action subsequence ($\alpha$) existing in a single plan or a set of plans ($\theta$). The support of a sequence $supp^\theta (\alpha)$ is the number of plans in $\theta$ that contain $\alpha$. Given a support threshold $\sigma$, a sequence $\alpha$ is a \textit{frequent sequence} on $\theta$ if $supp^\theta (\alpha) \geq \sigma$.  Mining of sequential patterns consists of finding the complete set of frequent subsequences given a set of plans $\theta$ and support threshold $\sigma$. 

Our method to learn and to use macro actions includes two algorithms. The \textit{set-up algorithm} takes as input a set of non-empty solution plans. It obtains a set of closed sequential patterns by using the BIDE+\cite{bideplus} algorithm. A \textit{closed sequential pattern} is a sequential pattern such that it is not strictly included in another pattern having the same support. Afterwards, each obtained sequence is evaluated regarding if each one of its actions belongs to the encoded operators of the problem. When it occurs the whole sequence is encoded and added in the problem macro-action list.
The \textit{enhanced search algorithm} uses the encoded macro-actions list to speed-up the search of a classical implementation of the A* algorithm.  A node $x$ is selected from the pending nodes list taking into account its heuristic value $h$. A plan is reached when $x$  satisfy the goal. If not the applicability of each  macro-action $m_i = \langle a_1a_2...a_n\rangle$  over $x$ is evaluated . A macro-action is applicable in $x$  when the preconditions of $x$ satisfies the preconditions of $m_i[0]$ allowing to get the successor $x'$ and for each obtained successor  the next actions are applicable ( from $n>0, m_i[n-1]$ is applicable to $x^{n-1}$ ). The created successors update the list of pending nodes and the list of explored nodes. After the algorithm try to apply the problem operators. Finally, the node $x$ is added to the explored nodes and another node is selected from the pending nodes list. 

\end{sloppypar}

\section{Results}
\begin{sloppypar}
The experiments were based on barman, depots, ferry and sokoban benchmarks. They were carried out on an Intel Xeon E5-2630 2.30GHz. The allocated CPU time was set to 300 seconds with a maximum of 8GB of memory. For each benchmark, a learning set of plans of 1000 problems and a test set of 300 problems were randomly generated with the generators\footnote{https://bitbucket.org/planning-tools/pddl-generators} used for the International Planning Competition (IPC).  We  went trough the SPMF \cite{JMLR:v15:fournierviger14a} data mining library, which implements the BIDE+ algorithm, to get the set of frequent sequences varying the degree of support between 10\% and 30\%. We used the PDDL4J \footnote {https://github.com/pellierd/pddl4j} library to encode them into a forward chaining planner based on A* algorithm and on FF heuristic. 

The evaluation was based on the classical metrics of quality and time used in IPC. Time score is computed as the quotient $T*/T$ where $T*$ is the minimum time required by the planner to solve the problem, and $T$ is the time spent by the evaluated implementation to solve the same planning task. Quality score is computed as $Q*  /Q$ where $Q*$ is the cost of the best known plan for a particular problem and Q is the cost of the plan produced by the evaluated implementation. If the planner found no solution the quality is set to zero. Table\ref{Table:Evaluation} shows for the implementation of each given support of frequent sequences as macro-actions, the improvement percentage in regard to the original algorithm.


\begin{table}[h!]
\centering
\begin{tabular}{l r r r r r r r r} 
 \hline
 Domain & \multicolumn{3}{c}{Time} & \multicolumn{3}{c}{Quality} \\ [0.5ex]
 \hline\hline
 {}   & supp10    & supp20   & supp30 & supp10    & supp20   & supp30 \\
 Barman & 108\% & 12\% & 2\% &32\% & -2,2\% &-0,8\%\\ 
 Depots & 25,3\%& 14,3\%& 4,8\%&0,6\%&1,4\%&0,5\%\\
 Ferry & 20,8\%&-28\%	&-7,6\%&-10\%&-2,6\% &0\%\\
 Sokoban & 172,1\%&113,9\%&	72,1\%&	-6,1\%&-0,8\%&-0,2\%\\ [1ex]
 \hline
\end{tabular}
\caption{Improvement percentage}
\label{Table:Evaluation}
\end{table}

Macro-actions quality score was improved in some domains owing the fact that the original algorithm could not found a plan for some problems (barman, depots and ferry). However, a domain where all the problems were solved by the original algorithm shows the worst quality (sokoban).  In our results, time improvement is obtained with a support of 10\% for all the domains. The results suggest that: (1) it can be possible to identify potential macro-actions over a domain by fixing the degree of support between 10\% and 30\%; (2) search performance can be improved, thus validating the relevance of macro-actions learning in the planning search.
Further work include a formalization of useful macro-actions as macro-operators based on the generalization of frequent sequences in order to deal with the utility problem, i.e., the increasing of the branching factor due to the adding of macro-actions. As well as, the development of a solution to learn online macro-actions. 
\end{sloppypar}

%
%
\bibliography{biblio}
\end{document}